\documentclass[12pt]{article}
\usepackage{graphicx}
\usepackage{titlesec}
\usepackage{amsmath,amssymb,bm,subfigure,color,url,booktabs,mathrsfs}%cite
\usepackage[colorlinks=true,citecolor=amazonite,linkcolor=amazonite,urlcolor=amazonite]{hyperref}
\usepackage{cite}

\definecolor{amazonite}{RGB}{0,115,150}
\definecolor{myred}{RGB}{255,56,0}
\definecolor{mygreen}{RGB}{30,150,30}
\definecolor{mybrown}{RGB}{150,30,30}
\definecolor{darkblue}{RGB}{30,50,100}
\definecolor{darkred}{RGB}{80,30,30}

\titleformat*{\section}{\color{darkred}\centering\large\bfseries}
\titleformat*{\subsection}{\color{darkred}\bfseries}
\titleformat*{\subsubsection}{\color{darkred}\it}
\titleformat*{\paragraph}{\color{darkred}\it}

\begin{document}

\renewcommand{\thefootnote}{\fnsymbol{footnote}}

{
\centering
{\color{darkred}\large\bf Statistical Learning and Estimation of Piano Fingering\footnote[1]{This work is in part supported by JSPS KAKENHI Nos.\ 16H01744, 16H02917, 16K00501, 16J05486, 17H00749, and 19K20340 and JST ACCEL No.\ JPMJAC1602.}}\\

\vskip 0.4cm
Eita Nakamura$^{1,2,}$\footnote[2]{Corresponding author. Electronic address: \tt{enakamura@sap.ist.i.kyoto-u.ac.jp}}, Yasuyuki Saito$^3$, and Kazuyoshi Yoshii$^{1}$
\vskip 0.2cm

{\it\small
$^1$Graduate School of Informatics, Kyoto University, Kyoto 606-8501, Japan\\
$^2$The Hakubi Center for Advanced Research, Kyoto University, Kyoto 606-8501, Japan\\
$^3$Department of Information and Computer Engineering, National Institute of Technology, Kisarazu College, Chiba 292-0041, Japan\\
}
\vskip 0.7cm
}

\begin{abstract}
Automatic estimation of piano fingering is important for understanding the computational process of music performance and applicable to performance assistance and education systems. While a natural way to formulate the quality of fingerings is to construct models of the constraints/costs of performance, it is generally difficult to find appropriate parameter values for these models. Here we study an alternative data-driven approach based on statistical modeling in which the appropriateness of a given fingering is described by probabilities. Specifically, we construct two types of hidden Markov models (HMMs) and their higher-order extensions. We also study deep neural network (DNN)-based methods for comparison. Using a newly released dataset of fingering annotations, we conduct systematic evaluations of these models as well as a representative constraint-based method. We find that the methods based on high-order HMMs outperform the other methods in terms of estimation accuracies. We also quantitatively study individual difference of fingering and propose evaluation measures that can be used with multiple ground truth data. We conclude that the HMM-based methods are currently state of the art and generate acceptable fingerings in most parts and that they have certain limitations such as ignorance of phrase boundaries and interdependence of the two hands.
\end{abstract}

\vspace{-2mm}
\vspace{-2mm}

\renewcommand{\thefootnote}{\arabic{footnote}}

%%%%%%%%%%%%%%%%%%%%%%%%%%%%%%%%%%%%%%%%%%%%%%
\section{Introduction}
\label{sec:Intro}
%%%%%%%%%%%%%%%%%%%%%%%%%%%%%%%%%%%%%%%%%%%%%%

Planning and coordination of relevant movements are essential for realizing music performance, which is one of the most skilled movements of humans \cite{Furuya2011,Palmer1997}.
For keyboard and string instruments, acquisition of appropriate fingering is considered a basic performance skill \cite{CPEBach,Couperin,Mozart,Musafia1971,Sor,Turk}.
From a computational viewpoint, finding an appropriate fingering for these musical instruments often involves a complex combinatorial optimization problem.
Automatic fingering estimation \cite{Kasimi2007,Balliauw2017,Prisco2012,Hart2000,Jacobs2001,Miura2004,Nagata2014,Nakamura2014,Parncutt1997,Tuohy2005,Yonebayashi2007} has been a topic of music information processing with the aim of understanding the computational process of music performance and is used both in performance assistance and education systems \cite{Lin2006,Saito2016,Sebastien2012,Takegawa2006} as well as for music arrangement \cite{Hori2013,Nakamura2018}.
There are also a growing number of public repositories\footnote{\url{https://github.com/search?q=piano+fingering&type=Repositories}} related to piano fingering generation, indicating the general interests in this topic.
Here we study methods for estimating piano fingering that are based on statistical modeling.

The quality of a fingering is thought to be determined on the bases of ergonomic, cognitive, and music-interpretive constraints \cite{Parncutt1997}, and previous studies have focused on the ergonomic constraints or costs of performance \cite{Kasimi2007,Balliauw2017,Hart2000,Parncutt1997}.
However, it is generally difficult to find, either theoretically or empirically, appropriate values for model parameters, i.e. the weights for constraints or the parameters of the cost function.
In an alternative approach based on statistical models \cite{Nakamura2014,Yonebayashi2007}, the constraints and costs of fingering are incorporated in these models through probabilities of fingering events.
This approach has the advantage that the parameters can be optimized on the basis of statistical learning from data.
Another approach is to use deep neural networks (DNNs) \cite{Goodfellow2016} that can learn nonlinear maps between input data (musical sequences) and output data (fingering).
DNNs can also be trained in a data-driven way and they can be used as universal approximators: given a sufficient number of layers and/or units and an appropriate parametrization, they can approximate a very wide class of maps arbitrarily well (see e.g.\ Section 6.4.1 in \cite{Goodfellow2016}).
Indeed, DNNs have been successfully applied to various pattern recognition problems.
To our knowledge, however, DNN-based methods for piano fingering estimation have not been reported in the literature.

To apply statistical learning, deep learning, or other data-driven methods for piano fingering, we need annotated data of musical note sequences with corresponding fingerings.
They are also necessary for systematically evaluating fingering estimation methods.
However, there has been no large-scale dataset publicly available and no large-scale evaluations of fingering estimation methods have been reported previously.

In this paper, we present a newly released dataset of piano fingering and propose fingering estimation methods based on hidden Markov models (HMMs) and DNNs.
The HMM \cite{Rabiner1989} is a basic statistical model for describing sequential dependence of latent variables (here, finger numbers) and correspondence between these variables and observed data (here, musical note sequences).
It has been applied to piano fingering estimation \cite{Nakamura2014,Yonebayashi2007}.
We propose higher-order extensions of previously studied HMMs and a refinement for incorporating constraints for polyphonic music.
We also propose another HMM (called chord HMM) with a distinct architecture where the latent states are defined at the chord level, instead of the note level as in the previous models.
This model can be considered as a probabilistic formulation of the cost-based model of \cite{Kasimi2007}.
As DNN-based methods, we study the feed-forward (FF) network and the long short-term memory (LSTM) network, which are commonly used in other domains such as speech recognition and machine translation.
Since there can be multiple appropriate fingerings that depend on the piano player's hand shape, knowledge, musical expressions, etc., we conduct statistical analyses of the individual difference of piano fingering.
Finally, we conduct comparative evaluations of the studied models and a representative constraint-based method \cite{Balliauw2017}.

Our main results are as follows.
\begin{itemize}%\setlength{\itemindent}{-3pt}
\item Construction of a large-scale piano fingering dataset (called PIG Dataset) including fingerings by multiple pianists, which is to our knowledge the first public dataset of the kind.
\item Quantitative measurement and characterization of individual differences in piano fingerings by multiple players.
\item Proposal of new methods for fingering estimation. The high-order HMMs achieved the state-of-the-art performance. The chord HMM and DNN-based methods have their own advantages and could be useful in the future.
\item Novel evaluation measures for comparing estimated fingerings with multiple ground truths.
\item Systematic evaluations that analyze the state-of-the-art performance, the effect of different architectures of the HMMs, the influence of training data size, typical estimation errors, and current limitations.
\end{itemize}

We review existing methods for piano fingering estimation in Section~\ref{sec:RelatedWork}.
We describe the dataset in Section~\ref{sec:Dataset}.
Statistical analyses on the individual difference of piano fingering using the dataset are presented in Section~\ref{sec:DataAnalysis}.
Methods for piano fingering estimation based on statistical models and DNNs are described in Section~\ref{sec:Model}, and the evaluation results are presented in Section~\ref{sec:Evaluation}.
We conclude in Section~\ref{sec:Concl}.

%%%%%%%%%%%%%%%%%%%%%%%%%%%%%%%%%%%%%%%%%%%%%%
\section{Review of Previous Methods for Piano Fingering Estimation}
\label{sec:RelatedWork}
%%%%%%%%%%%%%%%%%%%%%%%%%%%%%%%%%%%%%%%%%%%%%%

Parncutt et al.~\cite{Parncutt1997} proposed a constraint-based method, which is one of the earliest computational methods for estimating piano fingerings.
The method was tested on a small number of short monophonic passages that have multiple possibilities of fingerings and it was found that fingerings by human players often match with fingerings found by the method as one with the least difficulty.
A refined version was later proposed by introducing a better pitch representation and additional constraints \cite{Jacobs2001}.
Lin et al.~\cite{Lin2006} studied an online extension of the method with an application for performance assistance.

Recently, Balliauw et al.~\cite{Balliauw2017} further extended the constraint-based method of \cite{Parncutt1997} to treat polyphonic passages.
They also proposed a variable neighborhood search (VNS) algorithm that can efficiently find the optimal fingering according to the considered constraints.
Although they demonstrated the efficacy of the model with a few example results, they did not report any systematic quantitative evaluation. 

Hart et al.~\cite{Hart2000} proposed a cost-based method for finding the optimal fingering for monophonic pieces.
They formulated the cost of fingering as the sum of local costs for playing a pitch pair defined for each finger pair.
An algorithm based on dynamic programming was developed to find the optimal fingering for a given pitch sequence.
De Prisco et al.~\cite{Prisco2012} studied a method based on the genetic algorithm.
The fitness function, corresponding to the fingering cost, is learned from data.
Although they reported a high accuracy ($89.2\%$), the method was only evaluated on two musical pieces and more information (lengths of the pieces, generated results, etc.) was not provided.
A cost-based method that is applicable to polyphonic pieces was proposed by Al Kasimi et al.~\cite{Kasimi2007}.
The model consists of ``vertical costs'' representing the spread of the fingers involved in playing a particular chord and ``horizontal costs'' representing the stretch in the hand between two notes in consecutive chords (see Section~\ref{sec:ChordHMM} for more details), which can be seen as a generalization of the model of \cite{Hart2000}.

The first method for piano fingering estimation based on statistical modeling was proposed by Yonebayashi et al.~\cite{Yonebayashi2007}.
They proposed an HMM whose transition probabilities represent the statistical tendency of fingering motion and whose output probabilities represent that of hand shape.
Nakamura et al.~\cite{Nakamura2014} extended the HMM-based method to deal with fingering for both hands.
Their HMM can also be applied to polyphonic pieces and has a refined architecture of the output probabilities (see Section~\ref{sec:SimpleHMM} for more details).

%%%%%%%%%%%%%%%%%%%%%%%%%%%%%%%%%%%%%%%%%%%%%%
\section{Piano Fingering Dataset (PIG Dataset)}
\label{sec:Dataset}
%%%%%%%%%%%%%%%%%%%%%%%%%%%%%%%%%%%%%%%%%%%%%%

%%%
\subsection{Summary of the Dataset}
%%%

%
\begin{table}[t]
\centering
\caption{The subsets of the fingering dataset.}
{\tabcolsep = 2pt
\begin{tabular}{cccc}\toprule
Subset & Composers & Pieces (bars; notes) & \begin{tabular}{c}Pieces (bars; notes)\\with different fingerings\end{tabular}\\
\midrule
Bach   & 1  & 10 (218; 3,657) & 40 (872; 14,628) \\
Mozart & 1  & 10 (185; 2,546)  & 60 (1,110; 15,276) \\
Chopin & 1  & 10 (244; 4,022)  & 50 (1,220; 20,110) \\
Miscellaneous & 24 & 120 (2,533; 38,501) & 159 (3,355; 50,030) \\
\midrule
All           & 24 & 150 (3,180; 48,726) & 309 (6,557; 100,044) \\
\bottomrule
\end{tabular}}
\label{tab:BriefDescriptionOfData}
\end{table}
{\it PIG Dataset} (PIano fingernG) consists of piano pieces by Western classical music composers with fingerings annotated by pianists.
Currently there are 150 pieces in it and fingerings by one or more pianists are given for each piece.
To increase the variety of included music styles, we annotate a section of each musical piece for the fingering data, instead of annotating a small number of whole pieces.
The typical length of an annotated section is one page, about 20 bars, and about 300 notes.

There are special subsets that are intended for studying individual difference in fingerings for each piece.
These consist of 10 pieces by J.~S.~Bach (Bach set), 10 pieces by W.~A.~Mozart (Mozart set), and 10 pieces by F.~Chopin (Chopin set).
Each piece in these subsets is given fingerings by at least four different pianists.
The set of the other pieces is called the miscellaneous subset and consists of 120 pieces by 24 composers.
The miscellaneous subset also contains pieces by Bach, Mozart, and Chopin, but these pieces are different from those contained in the composer-specific subsets.
The sizes of the subsets are summarized in Table \ref{tab:BriefDescriptionOfData}.
The complete list of musical pieces is available on the web\footnote{http://beam.kisarazu.ac.jp/research/PianoFingeringDataset/}.

Fingerings in the dataset were provided by experienced pianists who graduated from a music college or who had played the piano for more than twenty years.
The pianists were asked to choose pieces that they could play and provided the fingering that they had actually used for the performance.

%%%
\subsection{Data Format}
%%%

There are two types of fingering data in general.
The first one is a set of finger numbers indicated in sheet music and is used to suggest a fingering (or fingerings) for the piece.
The other type is a record of finger numbers that are actually used for a performance by a particular player.
Our dataset contains the latter type of fingering data.
Since a piano performance can be represented as a MIDI signal where each note is described as a pair of note-on (key press) and note-off (key release) events, a fingering in our data is indicated by assigning a finger number to each MIDI note.

A PDF file of the musical score and a {\it fingering file} are provided for each fingering in the dataset.
In the fingering file, each note is described as a line in the order of onset time.
Each line has the following data format:
\begin{equation*}
n~~~t_{\rm on}~~~t_{\rm off}~~~p~~~v_{\rm on}~~~v_{\rm off}~~~c~~~f
\end{equation*}
where $n$ is the note ID, $t_{\rm on}$ and $t_{\rm off}$ are onset and offset times in seconds, $p$ is the pitch, $v_{\rm on}$ and $v_{\rm off}$ are onset and offset velocities (note intensity), $c$ indicates the hand part ($0=$ right hand (RH) and $1=$ left hand (LH)), and $f$ is a finger number ($1 = \text{thumb}, 2 = \text{index finger}, \cdots, 5 = \text{little finger}$ on the right hand and negative values represent corresponding fingers on the left hand).
Finger substitutions are notated like \verb$1_2$ (substitution from the right thumb to the right index finger).
From a fingering file one can create a MIDI signal for the piano performance.

%%%
\subsection{Data Preparation}
%%%

The musical pieces were selected by the present authors.
We intended to include frequently performed pieces by commonly known composers and to cover a wide range of music styles and time periods.
We also chose pieces whose digital-format musical scores were easy to obtain.
After selecting a section of each piece for labelling, a digital-format musical score file was prepared for each piece.
From the score file, a MIDI file and a PDF file were generated.
The content of the MIDI file is used for creating a fingering file.

Fingerings were first given by pianists as hand-written indications on printed musical scores and later they were transcribed into a digital format.
In the transcription process we double checked that fingerings were correctly formatted and if a dubious fingering was found we asked the pianist to double check it.

%%%%%%%%%%%%%%%%%%%%%%%%%%%%%%%%%%%%%%%%%%%%%%
\section{Statistical Analysis of Individual Difference}
\label{sec:DataAnalysis}
%%%%%%%%%%%%%%%%%%%%%%%%%%%%%%%%%%%%%%%%%%%%%%

A major characteristic of the dataset is that there are multiple fingerings for some portion of pieces, especially in the composer-specific subsets.
Here we conduct statistical analyses on these subsets to quantitatively study the nature of individual difference in piano fingering.
The results will be important in evaluating fingering estimation methods (see Section~\ref{sec:Evaluation}).
In the following, for simplicity, finger substitutions are neglected for analyses, modelling, and experiments: when we find a finger substitution in the data\footnote{Finger substitutions appear for only $0.34\%$ of the notes in the whole PIG Dataset.}, we use for analysis only the finger firstly used for pressing the piano key.

\begin{figure}[t]
\centering
{\includegraphics[clip,width=0.7\columnwidth]{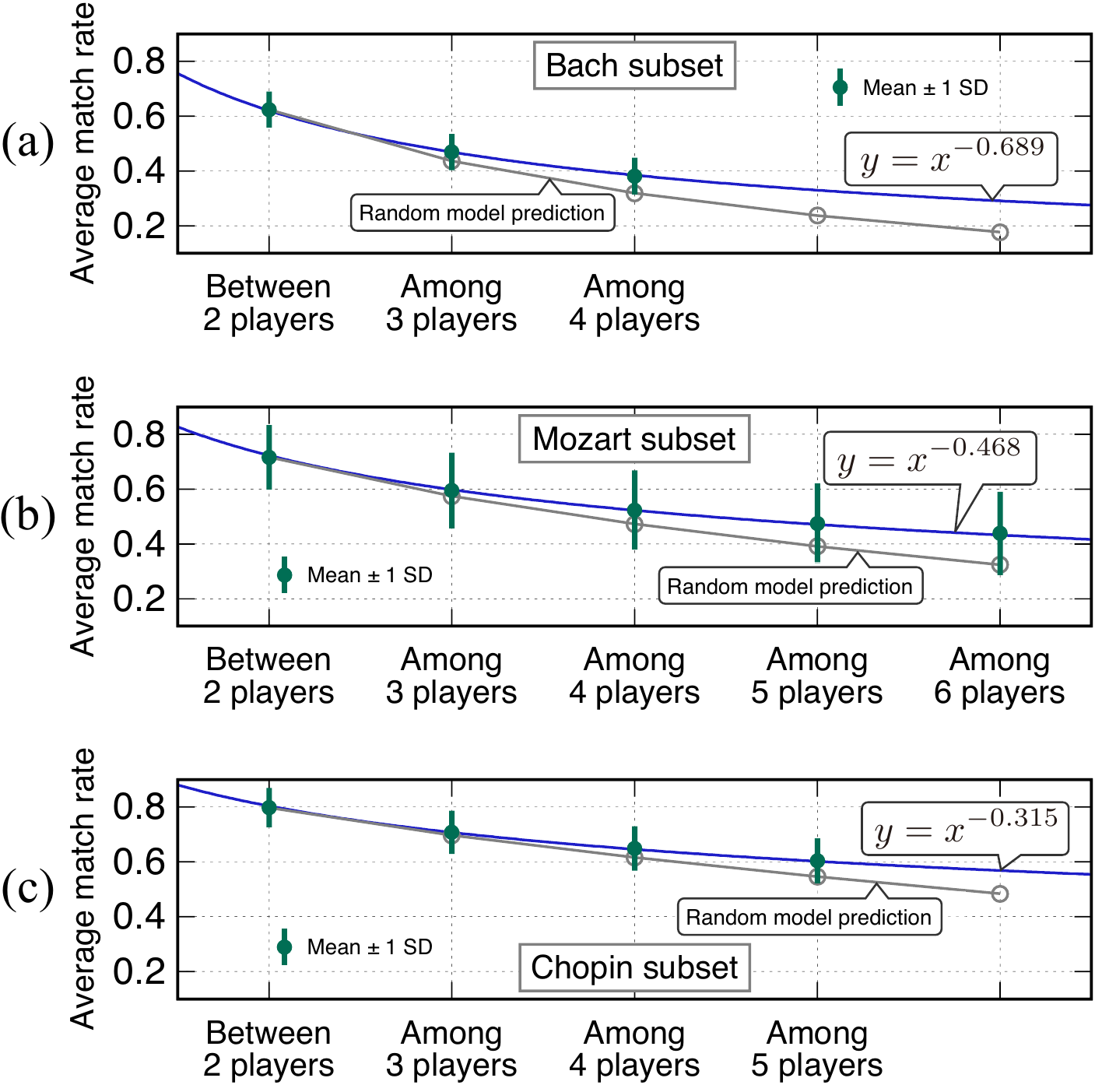}}
\caption{Average match rates of fingerings by different players. For each subset with fingerings of $J$ players, match rates for all combinations of $j$ ($2\leq j\leq J$) players were calculated and averaged. Predictions by the random model for $j\geq3$ were calculated from the values for $j=2$, assuming that fingerings are generated stochastically and independently for each note (see footnote \ref{foot:rand}).}
\label{fig:MatchRates}
\end{figure}
We first analyze how much fingerings by different players match.
To measure this, we use the {\it match rate} of a set of two or more fingerings for each piece, which is defined as the number of notes that are assigned the same finger divided by the total number of notes.
As there are four or more fingerings for each piece in the composer-specific subsets, we analyze the match rate between two players (averaged over all possible pairs of fingerings), the match rate among three players (averaged over all possible triples of fingerings), etc.\ up to the match rate among all the available fingerings.
The results in Fig.~\ref{fig:MatchRates} show that when comparing fingerings by two players, on average, $60\%$ to $80\%$ of the notes are played with the same fingers.

When comparing fingerings by three or more players, the proportion of notes that are played with an identical finger decreases but only slowly.
As shown in the figure, the average match rates follow a power function of the number of compared players.
In the figure, we also show reference values of match rates for three or more players that would be expected if fingerings are generated stochastically and independently for each note\footnote{\label{foot:rand}More precisely, these values are calculated by assuming that there are only two choices of fingers. In this case, the expected match rate for two sequences is given as $M_2=\omega^2+(1-\omega)^2$, where $\omega$ and $1-\omega$ are the probabilities for the two values. Once the value of $M_2$ is fixed, the value of $\omega$ can be solved. Then the expected match rate for $j$ sequences is given as $M_j=\omega^j+(1-\omega)^j$.}.
These expected values decrease exponentially with the number of compared players when it is large.
One can see that these upper bounds decrease faster than the real match rates.
This fact implies that the number of finger choices is different for individual notes and some notes have an almost unique finger choice.

\begin{figure}[t]
\centering
{\includegraphics[clip,width=0.6\columnwidth]{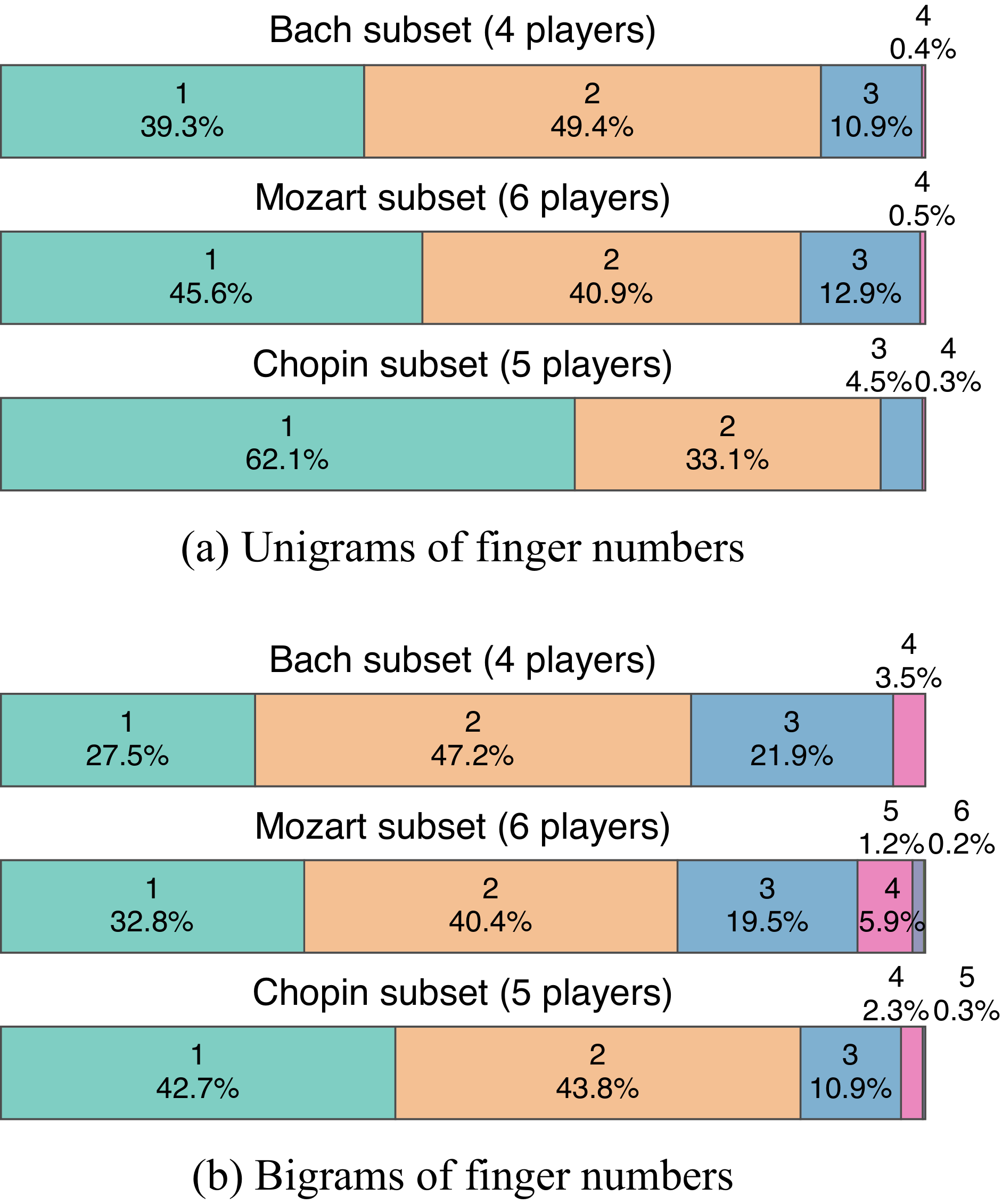}}
\caption{Multiplicities of the fingers used by different players. Proportions of the numbers of choices of finger(s) for each note/note-pair are shown.}
\label{fig:FingeringVariety}
\end{figure}
The distributions of the numbers of distinct fingers used for each note/note-pair in Fig.~\ref{fig:FingeringVariety} show the variety of fingerings from a local perspective.
For example, Fig.~\ref{fig:FingeringVariety}(a) shows that in the Bach subset the four players give only one choice of finger for $39.3\%$ of the notes and two choices for $49.4\%$ of the notes.
The results show that most notes have only one or two choices of finger and only a small portion of notes have three or four choices.
This shows that although a piano key can be played by any finger in principle, the choices of fingers actually used are severely limited and are determined by the context.
For example, the lower note of an octave chord in the RH part is almost always played with the thumb.
There is also a case that an apex note in a monophonic passage has an almost unique choice of finger.

Similarly, Fig.~\ref{fig:FingeringVariety}(b) shows the proportions of the numbers of choices of a pair of fingers for a pair of consecutive notes (in one hand part).
The results show that most note pairs have up to three choices of fingering.
Since a much larger proportion of four or more choices is expected if the choices of fingers for a note pair are made independently for each note, this indicates the strong sequential dependence on the choice of fingers.

Comparing the results for the three different composers, one finds that pieces by Bach have the smallest match rates and largest variety of finger choices, and that pieces by Chopin have the largest match rates and smallest variety of finger choices.
The main reason is probably that Chopin's pieces more frequently have larger chords, for which there are a smaller number of fingering choices.
The results may quantitatively demonstrate the general intuition that passages with denser notes have a lower degree of freedom for fingering, though we should note that the number of players providing fingerings is relatively small and differs among the subsets.

%%%%%%%%%%%%%%%%%%%%%%%%%%%%%%%%%%%%%%%%%%%%%%
\section{Models and Methods for Fingering Estimation}
\label{sec:Model}
%%%%%%%%%%%%%%%%%%%%%%%%%%%%%%%%%%%%%%%%%%%%%%

We review the previous method based on a simple HMM \cite{Nakamura2014,Yonebayashi2007} in Section~\ref{sec:SimpleHMM} and present the proposed higher-order extensions in Section~\ref{sec:HigherOrderHMMs}. In Sections~\ref{sec:ChordHMM} and \ref{sec:DNN} we describe proposed methods based on the chord HMM and DNNs, respectively.

%%%
\subsection{Problem Specification}
%%%

In this study, we consider the following specific problem of fingering estimation.
The input is a piano performance represented as a MIDI-like signal $(p_n,t_n,\bar{t}_n)_{n=1}^N$ where each musical note $n$ is described by its pitch $p_n$, onset time $t_n$, and offset time $\bar{t}_n$ ($N$ denotes the number of notes).
We assume that the notes are ordered according to their onset times.
We also assume that the LH and RH parts are separately given.
(For automatic separation of hand parts from a mixed signal of both hands, see for example \cite{Nakamura2014}.)
The output is a list of finger numbers $(f_n)_{n=1}^N$ corresponding to each note $n$.
Each $f_n$ can take a value in $\{1,\ldots,5\}$; they indicate $1 = \text{thumb}, 2 = \text{index finger}, \cdots, 5 = \text{little finger}$ as in the standard music notation.

%%%
\subsection{Simple HMM}
\label{sec:SimpleHMM}
%%%

Let us first explain the first-order HMM for fingering estimation, which has been proposed previously \cite{Nakamura2014,Yonebayashi2007}.

\subsubsection{Generative Model}

An HMM for piano fingering estimation describes the generative process of piano performances in a probabilistic manner.
The model provides a way to estimate the probability $P(\bm p,\bm f)$ of any pair of piano performance $\bm p=(p_n)_{n=1}^N$ and fingering $\bm f=(f_n)_{n=1}^N$.
In the context of HMM, pitches $\bm p$ are called observations or outputs and fingering numbers $\bm f$ are called (hidden) states.
First, the model generates the fingering according to a Markov model:
\begin{equation}
P(\bm f)=P(f_1)\prod_{n=2}^NP(f_n|f_{n-1}),
\label{eq:SimpleHMMTrProb}
\end{equation}
where $P(f_1)$ is called the initial probability and $P(f_n|f_{n-1})$ is called the (state) transition probability.
The transition probabilities statistically describe the seqential dependence of the order of fingers.
Next, the model generates the pitches $\bm p$ conditionally on the fingering $\bm f$ according to
\begin{equation}
P(\bm p|\bm f)=P(p_1|f_1)\prod_{n=2}^NP(p_n|\,p_{n-1},f_{n-1},f_n),
\label{eq:SimpleHMMOutProb}
\end{equation}
where the factors in the RH side are called output probabilities.
The probabilities $P(p_n|\,p_{n-1},f_{n-1},f_n)$ statistically describe the correspondence between a pair of pitches and a pair of finger numbers.
The first factor $P(p_1|f_1)$ is formally necessary to generate the pitch sequence.
The joint probability $P(\bm p,\bm f)=P(\bm f)P(\bm p|\bm f)$ can be obtained by the composition of the probabilities in Eqs.~(\ref{eq:SimpleHMMTrProb}) and (\ref{eq:SimpleHMMOutProb}).

In the above formulation, the model does not describe onset and offset times; the model cares about the temporal order of notes but not about their temporal distances.
Temporal distances of notes have rarely been considered in previous studies and they are not considered also in this study.
To describe the irregularity regarding chords (notes played almost simultaneously), however, we will present a model that partially use temporal information.
This model is described in Section~\ref{sec:DescriptionOnChords}.

\subsubsection{Design of Output Probabilities}
\label{sec:DesignOfOutProb}

In the most general setting, the output probabilities $P(p_n|\,p_{n-1},f_{n-1},f_n)$ have $88^2\cdot5^2(\sim 2\cdot 10^5)$ parameters.
Since it is difficult to reliably estimate all these parameters from a limited amount of data, it is necessary to introduce assumptions on the model structure that reduce the effective number of parameters.
There are variants of HMMs for fingering estimation that differ in the assumptions put on the output probabilities.

\begin{figure}[t]
\centering
{\includegraphics[clip,width=0.5\columnwidth]{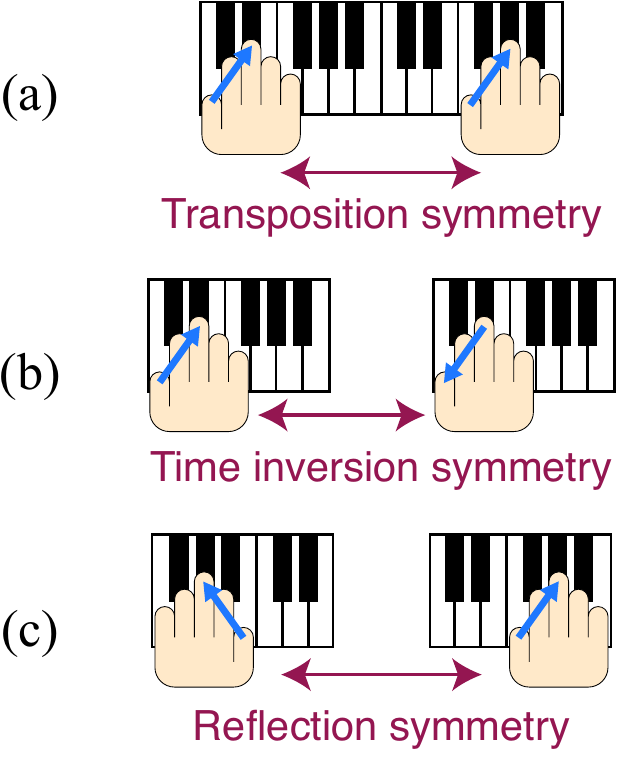}}
\caption{Symmetries of the output probabilities of the HMM.}
\label{fig:Symmetries}
\end{figure}
A common assumption is that the output probabilities depend on the relative pitch interval but not on absolute pitches.
This {\it transposition symmetry} (Fig.~\ref{fig:Symmetries}(a)) of the output probabilities is mathematically expressed as
\begin{equation}
P(p_n|\,p_{n-1},f_{n-1},f_n)=P(p_n+p\,|\,p_{n-1}+p,f_{n-1},f_n)
\end{equation}
for any pitch $p$, and consequently the probability distribution has the following form:
\begin{equation}
P(p_n|\,p_{n-1},f_{n-1},f_n)=F(p_n-p_{n-1};f_{n-1},f_n).
\label{eq:TranspositionSymOutProb}
\end{equation}
This assumption is applied throughout this paper.
Addition and subtraction operations for pitches, which are explained below, should be defined to write these equations.

\begin{figure}[t]
\centering
{\includegraphics[clip,width=0.5\columnwidth]{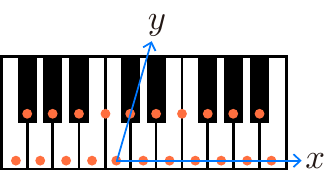}}
\caption{Piano keyboard represented as a two-dimensional lattice.}
\label{fig:KeyboardLattice}
\end{figure}

There are several possibilities for pitch representation, and in this paper we consider two typical representations:
\begin{itemize}\setlength{\itemindent}{-3pt}
\item Integral pitch representation
\item Lattice pitch representation
\end{itemize}
In the first representation, a pitch is represented as an integer in units of semitones.
It is mathematically simple and addition and subtraction operations can be defined straightforwardly.
A shortcoming of this representation is that it cannot reflect the geometric nature of the piano keyboard consisting of black and white keys, which is important for fingering \cite{Jacobs2001}.
To solve this problem, in the second representation, the piano keyboard is described as a two-dimensional lattice and a pitch is represented as a point in the lattice (Fig.~\ref{fig:KeyboardLattice}).
In this representation, addition and subtraction can be defined to act on the two dimensions independently.
For example, in the $y$-direction the value of $p_n-p_{n-1}$ can be either $0$, $1$, or $-1$.

Another basic assumption is to treat very large leaps equally.
Since leaps much larger than one octave appear only occasionally, it is impractical to precisely estimate probabilities for individual cases.
To deal with this problem, in the integral pitch representation we can introduce a cutoff $\delta p_{\rm max}$ and treat the cases with $p_n-p_{n-1}>\delta p_{\rm max}$ equally (and do something similar for the cases with $p_n-p_{n-1}<-\delta p_{\rm max}$).
In the lattice pitch representation, such a cutoff can be applied for the $x$-direction.

Two more assumptions are proposed in a previous study \cite{Nakamura2014}.
One is to impose {\it time inversion symmetry} (Fig.~\ref{fig:Symmetries}(b)), which is expressed as
\begin{equation}
F(p_n-p_{n-1};f_{n-1},f_n)=F(p_{n-1}-p_n;f_n,f_{n-1}).
\end{equation}
The equation says that a backward fingering motion costs as much as the forward motion, which is motivated by the underlying physical process.
The other assumption is that the model parameters for the LH and RH are related by reflections in the $x$-direction.
This {\it reflection symmetry} (Fig.~\ref{fig:Symmetries}(c)) can be expressed as
\begin{equation}
F_L(p_n-p_{n-1};f_{n-1},f_n)=F_R(-[p_n-p_{n-1}]_x;f_{n-1},f_n),
\end{equation}
where $F_L$ and $F_R$ are output probabilities for the LH and RH, and $-[\,\cdot\,]_x$ denotes a minus operation in the $x$-direction: it acts as the standard minus operation in the integral pitch representation and it only inverts the sign of the argument in the $x$-direction in the lattice pitch representation.
\subsubsection{Description of Chords}
\label{sec:DescriptionOnChords}

We have assumed that the input note sequence $(p_n,t_n,\bar{t}_n)_{n=1}^N$ is ordered according to onset time, but so far temporal information is not incorporated in the model.
Consequently, the model is already applicable for a general polyphonic piece where chords can be described by multiple notes with the same or almost the same onset times.
Without using temporal information, however, constraints specific to the fingering of chords (e.g.\ the rarity of finger crossing within a chord) cannot be imposed.

We can refine the model to take into account the fingering constraints specific for chords by introducing two sets of transition and output probabilities that are separately applied for note transitions within a chord and those across chords.
However, we have observed in a preliminary study that the effect of this refinement is limited because of the limited amount of available training data.
Therefore, here we consider a minimal model that incorporates the most important constraint: avoid finger crossing within a chord.
This can be realized by setting the output probability to zero when the direction of pitch transition and that of finger movement are opposite for note transitions within a chord.
Whether two notes are part of a chord or not can be detected by looking at the difference between their onset times.
Specifically, it is known that the condition $|t_n-t_{n-1}|\leq\Delta$ with $\Delta=30$ ms indicates that the $(n-1)$\,th and $n$\,th notes are part of a chord (Fig.~4 in \cite{Nakamura2014JNMR}).
In our refined model, the constraint on finger crossings is applied when this condition is satisfied.

\subsubsection{Fingering Estimation Algorithm}

In the statistical framework, fingering estimation can be formulated as a probabilistic optimization problem.
For a given sequence of notes $\bm p$, we obtain as an estimate of the optimal fingering the most probable fingering $\hat{\bm f}$ that maximizes $P(\bm f|\bm p)\propto P(\bm f,\bm p)$.
For the HMM considered in this study, we can apply the standard Viterbi algorithm \cite{Rabiner1989} to find this fingering efficiently.
Note that while the fingering is generated in a forward direction by the HMM, the optimal fingering is estimated by taking into account the whole sequence $\bm f$ of fingering and $\bm p$ of pitches including the previous and future events.

%%%
\subsection{Higher-Order HMMs}
\label{sec:HigherOrderHMMs}
%%%

The simple HMM in Section~\ref{sec:SimpleHMM} has the limitation that only the most local fingering constraints involving consecutive notes can be described.
This limitation can be overcome by higher-order HMMs, which in general have the following form of transition and output probabilities:
\begin{align}
&P(f_n|\,f_{n-m},\ldots,f_{n-1}),
\\
&P(p_n|\,p_{n-m},\ldots,p_{n-1},f_{n-m},\ldots,f_{n}),
\end{align}
where $m$ is the order of the HMM.

The output probability of an $m$\,th-order HMM in general has $88^m\cdot5^m$ parameters, which blows up quickly for increasing $m$.
A way to efficiently approximate the output probability with a small number of parameters is to decompose it into a set of pairwise factors as follows:
\begin{align}
P(p_n|\,p_{n-m},\ldots,p_{n-1},f_{n-m},\ldots,f_{n})
\propto\prod_{l=1}^mP(p_n|\,p_{n-l},f_{n-l},f_n)^{\alpha_l}.
\label{eq:PairwiseOutProb}
\end{align}
This equation says that the dependence of the previous pitches, current finger number, and previous finger numbers on the probability of the current pitch is given as a product of factors, each of which only involves a pair of the current note and one of the previous notes.
Such a pairwise decomposition has been shown to be effective for modeling sequential data in other domains \cite{Sakellariou2017,Schneidman2006}.
We have introduced coefficients $\alpha_l$ that represent weights for the $l$\,th factor.
A larger value of $\alpha_l$ increases the influence of the $l$\,th factor in the calculation of the output probabilities.
Each factor in the RH side is assumed to have transposition symmetry and be expressed as in Eq.~(\ref{eq:TranspositionSymOutProb}).
Further assumptions on the output probabilities explained in Section~\ref{sec:DesignOfOutProb} can similarly be applied for higher-order HMMs.

The number of parameters for the transition probabilities also grows exponentially with respect to the order $m$, though less severely than does the number of parameters for the output probabilities.
To avoid possible data sparseness, we introduce a smoothing method for the transition probabilities based on linear interpolation:
\begin{align}
&P(f_n|\,f_{n-m},\ldots,f_{n-1})
\notag\\
&~=(1-\lambda_1-\cdots-\lambda_{m-1})P_{\rm ML}(f_n|\,f_{n-m},\ldots,f_{n-1})
\notag\\
&\qquad+\sum_{l=1}^{m-1}\lambda_lP_{\rm ML}(f_n|\,f_{n-l},\ldots,f_{n-1}),
\label{eq:SmoothedTrProb}
\end{align}
where $\lambda_l\in[0,1]$ denote the interpolation coefficients and $P_{\rm ML}(f_n|\,f_{n-l},\ldots,f_{n-1})$ are the $l$\,th-order transition probabilities before being smoothed.
For example, even if we have $P_{\rm ML}(f_n|\,f_{n-m},\ldots,f_{n-1})=0$ for some configuration $(f_{n-m},\ldots,f_n)$ of finger numbers, the probability $P(f_n|\,f_{n-m},\ldots,f_{n-1})$ is nonzero given that $\lambda_1\neq0$ and $P_{\rm ML}(f_n|f_{n-1})\neq0$.
Thus, we can avoid the zero-frequency problem using lower-order transition probabilities.

Given training data of music performances with fingering annotations, the transition probabilities and each pairwise output probability $P(p_n|\,p_{n-l},f_{n-l},f_n)$ in Eq.~(\ref{eq:PairwiseOutProb}) can be estimated straightforwardly.
On the other hand, the coefficients $\alpha_l$ and $\lambda_l$ can be determined by an optimization process according to the accuracy of the fingering estimation (see Section~\ref{sec:Evaluation} for details).
The fingering estimation algorithm can be derived by simply extending the Viterbi algorithm.

%%%
\subsection{Chord HMM}
\label{sec:ChordHMM}
%%%

We here construct another HMM, one that is a probabilistic formulation of the model proposed in \cite{Kasimi2007} (we call this model {\it chord HMM}).
Instead of considering state transitions for each note as in the models in Secs.~\ref{sec:SimpleHMM} and \ref{sec:HigherOrderHMMs}, the states of this model are defined for each ``chord''.
Here, a ``chord'' is defined as a set of notes with almost simultaneous onsets and, specifically, we use a threshold $\Delta=30$ ms (as in Section~\ref{sec:DescriptionOnChords}) to cluster a note sequence into chords.
As explained in detail below, owing to this model structure, the chord HMM is advantageous in incorporating transitional dependence of chords, which corresponds to long-range dependence at the note level, as well as the influence of note durations.

The states of the model are constructed as all possible combinations of fingers for each chord.
Assuming that finger crossings are not used within a chord and component pitches are played by distinct fingers, there are $\binom{5}{K}$ possible fingerings for a chord with $K$ pitches.
A key feature of this model is that offset times are also considered and as long as a note persists it is included as a component pitch of succeeding chords.
Neglecting the possibility of finger substitutions, state transitions are constrained so that the same finger is used for each sustained note.

In the original model of \cite{Kasimi2007}, the vertical and horizontal costs are considered.
Here, the current and previous chords are denoted by $Q=\{Q_k\}_{k=1}^K$ (with $K$ pitches) and $Q'=\{Q'_k\}_{k=1}^{K'}$ (with $K'$ pitches), their component pitches are denoted by $Q_k$ ($k\in\{1,\ldots,K\}$) and $Q'_k$ ($k\in\{1,\ldots,K'\}$), and the corresponding finger numbers are denoted by $G=\{G_k\}_{k=1}^K$ and $G'_k=\{G'_k\}_{k=1}^{K'}$ ($G_k$ represents the finger for pitch $Q_k$ and $G'_k$ represents the finger for pitch $Q'_k$).
The vertical cost $V$ represents the spread of the fingers involved in playing a particular chord and is described as
\begin{align}
V(Q,G)=\sum_{\substack{k,k'=1\\ k\neq k'}}^Kv(Q_{k'},Q_k;G_{k'},G_k),
\label{eq:VerticalCost}
\end{align}
while the horizontal cost $H$ represents the stretch in the hand between two notes in consecutive chords and is described as
\begin{align}
H(Q',G';Q,G)=\sum_{k'=1}^{K'}\sum_{k=1}^Kh(Q'_{k'},Q_{k};G'_{k'},G_k).
\label{eq:HorizontalCost}
\end{align}
Here, $v(p',p;f',f)$ and $h(p',p;f',f)$ are the costs of playing a pitch pair $(p',p)$ with a finger pair $(f',f)$; the former is applied for a vertical pitch interval within a chord and the latter for a horizontal pitch interval across chords.
The sum of vertical and horizontal costs serves as the cost of state transitions.

In the probabilistic formulation, the vertical and horizontal costs are represented as products of pairwise transition and output probabilities, as in the equation for the output probabilities for a high-order Markov model (Eq.~(\ref{eq:PairwiseOutProb})).
The transition and output probabilities are described as
\begin{align}
&P(G|G')\propto \bigg[\prod_{k'=1}^{K'}\prod_{k=1}^KP(G_k|G'_{k'})^{\beta_1}\bigg]
\bigg[\prod_{\substack{k,k'=1\\ k\neq k'}}^KP(G_k|G_{k'})^{\beta_2}\bigg],
\label{eq:ChordLevelHMMTrProb}
\\
&P(Q|Q',G',G)\propto
\bigg[\prod_{k'=1}^{K'}\prod_{k=1}^KP(Q_{k}|Q'_{k'},G'_{k'},G_k)^{\gamma_1}\bigg]
\notag\\
&\qquad\qquad\qquad\qquad\cdot\bigg[\prod_{\substack{k,k'=1\\ k\neq k'}}^KP(Q_{k}|Q_{k'},G_{k'},G_k)^{\gamma_2}\bigg].
\label{eq:ChordLevelHMMOutProb}
\end{align}
For the component probabilities in Eq.~(\ref{eq:ChordLevelHMMOutProb}) we impose transposition symmetry to reduce the number of parameters and use the lattice pitch representation.
As we do for the output probabilities of HMMs, we can also impose time inversion symmetry and reflection symmetry.

In the training phase we can simply estimate the component probabilities by the maximum-likelihood method and in the inference phase a simple Viterbi algorithm can be applied.
The products in Eqs.~(\ref{eq:ChordLevelHMMTrProb}) and (\ref{eq:ChordLevelHMMOutProb}) involve all pairs of component pitches, which may give too much weight to chords with a large number of notes.
To reduce this possible bias, we introduce a factor $K^{-\zeta}$ ($\zeta\geq0$) to weight the contribution of a chord with $K$ pitches in the Viterbi updates.
The coefficients $\beta_1$, $\beta_2$, $\lambda_1$, $\lambda_2$, and $\zeta$ can be determined with the objective of optimizing the fingering estimation accuracy.
The chord HMM can be extended for higher-orders in the same manner as the HMMs in Section~\ref{sec:HigherOrderHMMs}.

%%%
\subsection{DNN-Based Methods}
\label{sec:DNN}
%%%

In light of the recent extensive application of deep learning, we consider as references for comparative evaluation two simple DNNs for piano fingering estimation; a feedforward (FF) network and a long short-term memory (LSTM) network.
For both networks, the input is a sequence of integer pitches and the output is the corresponding fingering numbers.
Each network was separately trained and tested for the LH and RH parts.
We tried different architectures and compared the accuracy using the test data described in Section~\ref{sec:Setup}.
For both FF and LSTM networks, we tried the numbers of hidden layers $1$, $3$, and $5$, and the numbers of units $16$, $32$, and $64$.
We chose a $3$-hidden-layer FF network and a $3$-layer LSTM network with $32$ units and one softmax layer according to this optimization, but we also observed that the accuracy was not so sensitive to those hyperparameters.
The window size was $\pm5$ around the symbol of interest.

%%%%%%%%%%%%%%%%%%%%%%%%%%%%%%%%%%%%%%%%%%%%%%
\section{Evaluation}
\label{sec:Evaluation}
%%%%%%%%%%%%%%%%%%%%%%%%%%%%%%%%%%%%%%%%%%%%%%

%%%
\subsection{Setup}
\label{sec:Setup}
%%%

To systematically compare the performances of the piano fingering methods described in Section~\ref{sec:Model}, we trained and tested the methods using PIG Dataset explained in Section~\ref{sec:Dataset}.
We used the miscellaneous subset for training and used the three composer-specific subsets for testing.
In the testing stage, we used the musical notes in the upper and lower staves as the RH and LH parts, respectively.

We implemented the HMMs in Secs.~\ref{sec:SimpleHMM} and \ref{sec:HigherOrderHMMs} up to the third order, the first-order chord HMM in Section~\ref{sec:ChordHMM}, and the two DNN-based methods (FF and LSTM) in Section~\ref{sec:DNN}.
The source code for the HMM-based methods (as well as the tools for computing the evaluation measures described in Section \ref{sec:EvaluationMeasure}) is available in the accompanying webpage\footnote{\url{https://statpianofingering.github.io/demo.html}}.
The weight coefficients $\alpha_l$ for the HMMs and $\beta_1$, $\beta_2$, $\gamma_1$, $\gamma_2$, and $\zeta$ for the chord HMM were optimized by the Bayesian optimization method \cite{Shahriari2015}, which can optimize multi-dimensional parameters for a general objective function without requiring the gradients.
Specifically, we used the Gaussian process mutual information algorithm proposed in \cite{Contal2014} and the number of iterations was $200$.
Since parameters $\alpha_l$ have the effect of controlling the influence of the output probabilities relative to the transition probabilities, we also introduced a corresponding coefficient $\alpha_1$ for the first-order HMM and optimized it similarly.
$\delta p_{\rm max}$ was $15$.
For training the DNNs, we used the Adam optimizer \cite{Adam} and a batch size of $3$.

Among the methods proposed in previous studies reviewed in Section~\ref{sec:RelatedWork}, only the VNS method \cite{Balliauw2017} has public source code provided.
This method, which is a refinement of \cite{Parncutt1997}, can be considered as a representative constraint-based method.
Since the provided program takes a MusicXML file as input, we generated MusicXML files for the tested musical pieces and merged the obtained outputs into the form of our fingering data.
The default parameter values of the method were used for the experiment.

%%%
\subsection{Evaluation Measures}
\label{sec:EvaluationMeasure}
%%%

Considering that appropriate piano fingerings are not unique, evaluation measures that can be used for multiple ground truth data are necessary.
Since it is difficult to find one perfect evaluation measure, we here propose several possible measures.
When only one ground truth fingering is given, the simplest measure is the match rate, which is the fraction of notes for which the estimated fingers are correct.
When multiple ground truths are given, we can calculate the match rate for each ground truth and use their average value as the {\it general match rate} $M_{\rm gen}$ indicating how closely the estimation agrees with all the ground truths.
We can also focus on the ground truth closest to the estimation and define the {\it highest match rate} $M_{\rm high}$.
On the other hand, the softest criterion of correct estimation for each note is to judge whether the estimated finger matches at least one of the ground truths.
In this way, we can define the {\it soft-match rate} $M_{\rm soft}$.

\begin{figure}[t]
\centering
{\includegraphics[clip,width=0.7\columnwidth]{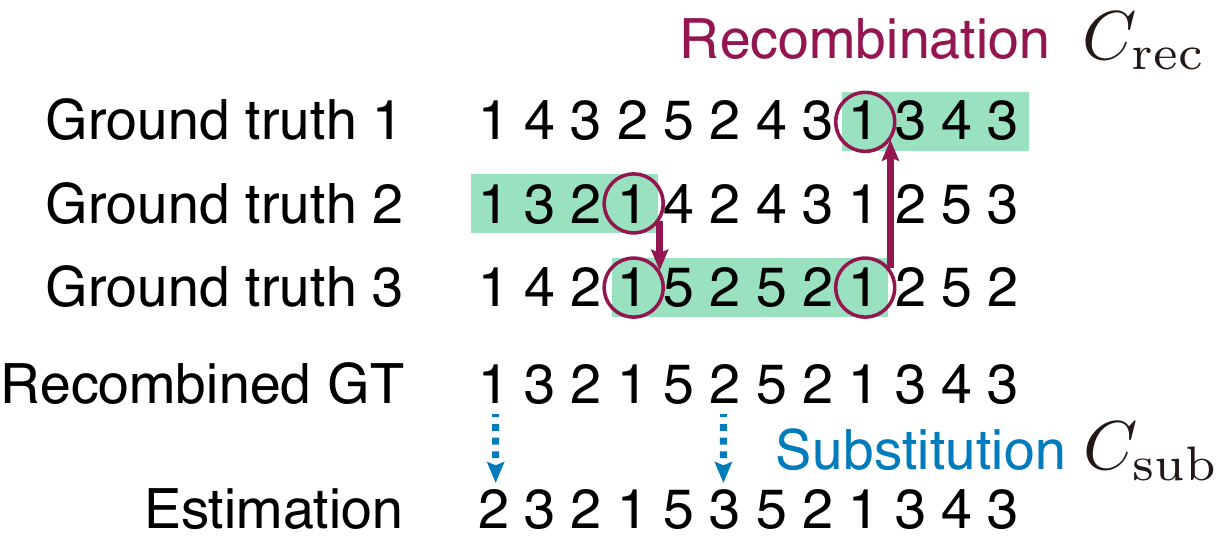}}
\caption{Edit costs for the recombination match rate.}
\label{fig:MultiGTMatchRate}
\end{figure}
As discussed in Section~\ref{sec:DataAnalysis}, there can be several choices of fingers locally and at the same time there is strong sequential dependence that rules out arbitrary finger transitions between these choices.
To accommodate this nature, we consider a measure that compares the estimation with a reference sequence that is constructed by recombining the multiple ground truths (Fig.~\ref{fig:MultiGTMatchRate}).
We formulate this measure based on an edit distance.
To construct this reference sequence (called {\it recombined ground truth}), we allow transitions between ground truths only at locations where the finger numbers match and assign an edit cost ({\it recombination cost}) $C_{\rm rec}$ for each transition.
We also assign an edit cost ({\it substitution cost}) $C_{\rm sub}=1$ for each mismatch of finger numbers between the estimation and the recombined ground truth.
The total edit cost $E_{\rm rec}$ for changing the multiple ground truths to the estimated sequence is interpreted as the amount of error.
We consider the recombined ground truth that minimizes the error of the estimation and define the {\it recombination match rate} $M_{\rm rec}=(N-E_{\rm rec})/N$, where $N$ is the total number of notes.
In the following, we consider $C_{\rm rec}=1$; that is, a recombination is penalized the same amount as a local mismatch.
Details of the algorithm for computing the recombination match rate are explained in \ref{app:RecombMatchRate}.

Evaluating the similarity between fingerings and the appropriateness of a given fingering is still an open problem and we do not think that the aforementioned measures capture all important aspects.
For example, the above match rates are insensitive to the quality of errors and do not distinguish “acceptable” and “unacceptable” errors.
It is also important to measure the perceptual quality of fingerings by subjective experiments by pianists, for which methodology for dealing with sequential data should be developed. 
These points need to be explored further in future research.

%%%
\subsection{Accuracy Comparisons}
%%%

\begin{figure}[t]
\centering
{\includegraphics[clip,width=0.75\columnwidth]{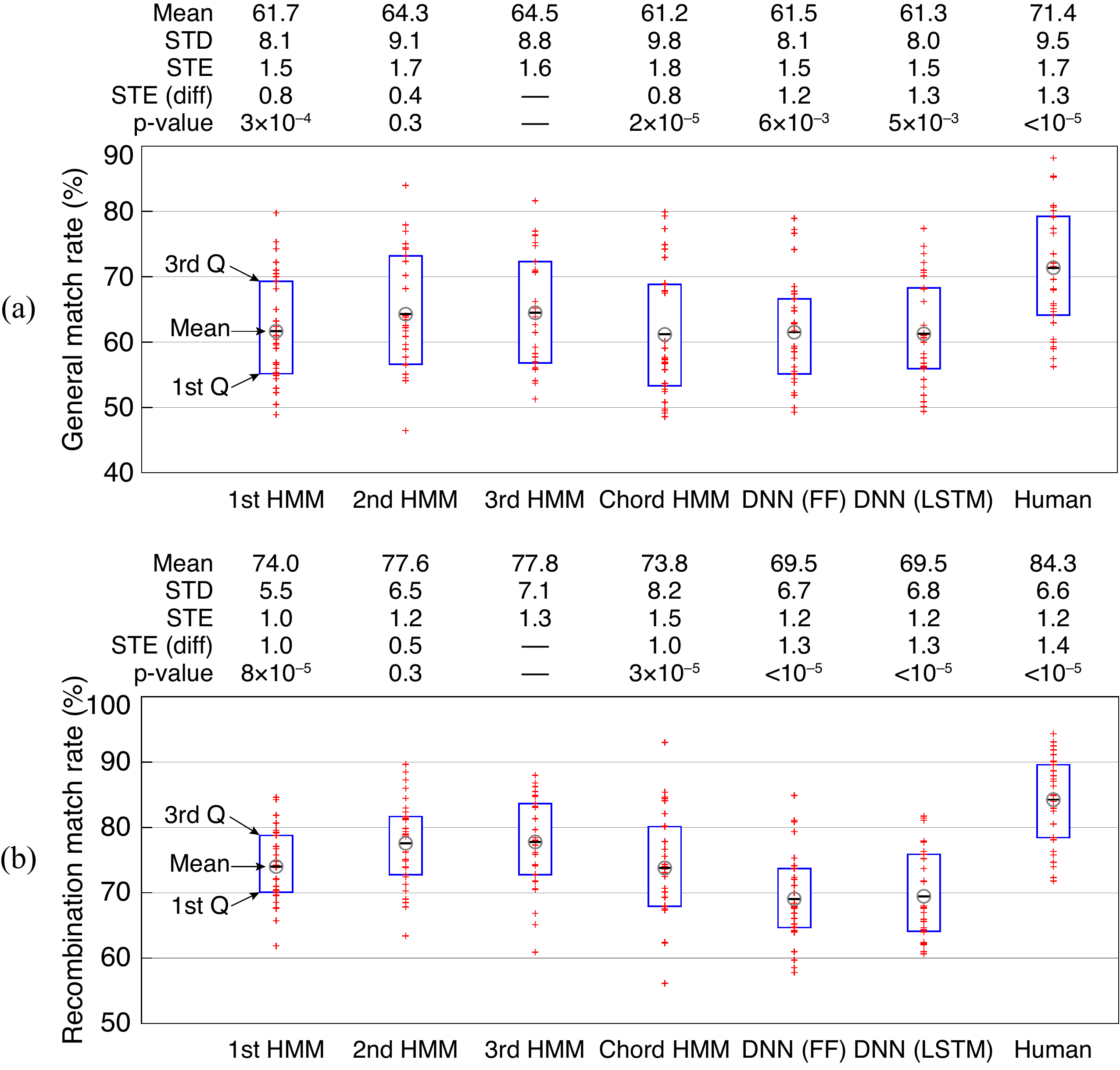}}
\caption{Detailed statistics of the general and recombination match rates on the test data. The list of values in each panel indicate, from top to bottom, the mean value, standard deviation, standard error, standard error of differences from the results of the third-order HMM, and p-value using t-test statistics for the difference of the mean value from the result of the third-order HMM. In the panels, circles indicate the mean value and boxes indicate the first and third quartiles.}
\label{fig:StatMR}
\end{figure}

%
\begin{table}[t]
\centering
\caption{Comparative evaluation of piano fingering estimation methods. The match rates are averaged over the 30 pieces in the test data. $^*$For the reference values (`Human'), the match rates for fingerings by each player in the test data are computed by comparing with all of the other annotators' ground truths.}
{\tabcolsep = 3pt
\begin{tabular}{rcccc}\toprule
Method & $M_{\rm gen}$ & $M_{\rm high}$ & $M_{\rm soft}$ & $M_{\rm rec}$\\
\midrule
1st HMM   & $61.7$ & $68.3$ & $82.8$ & $74.0$ \\
2nd HMM   & $64.3$ & $70.8$ & $85.3$ & $77.6$ \\
3rd HMM   & $64.5$ & $71.0$ & $85.5$ & $77.8$ \\
Chord HMM & $61.2$ & $67.7$ & $81.7$ & $73.8$ \\
DNN (FF)  & $61.5$ & $66.2$ & $82.5$ & $69.5$ \\
DNN (LSTM)& $61.3$ & $66.1$ & $82.8$ & $69.5$ \\
Human$^*$ & $71.4$ & $79.1$ & $90.8$ & $84.3$ \\
\bottomrule
\end{tabular}
}
\label{tab:ComparativeEvaluation}
\end{table}
Let us first discuss the results of comparison among all the methods (Table \ref{tab:ComparativeEvaluation} and Fig.~\ref{fig:StatMR}).
For the output probabilities of the HMMs and the chord HMM, only the transposition symmetry is imposed.
The values of the coefficients of these models used to obtain the results are provided in Table \ref{tab:OptimizedParameters}.
The constraint for chords (Section~\ref{sec:DescriptionOnChords}) is applied in the HMMs.
The result for the VNS method is not shown because the method failed to output fingerings for part of the test data; the comparison with this method is discussed later.

\begin{table}[t]
\centering
\caption{Parameters used to obtain the results in Table \ref{tab:ComparativeEvaluation}.}
{\tabcolsep = 3pt
\begin{tabular}{rl}\toprule
Model & Parameters\\
\midrule
1st HMM   & $\alpha_1=0.964$ \\
2nd HMM   & $\alpha_1=0.556$, $\alpha_2=0.407$, $\lambda_1=0.474$\\
3rd HMM   & \hspace{-3pt}\begin{tabular}{l}$\alpha_1=0.448$, $\alpha_2=0.292$, $\alpha_3=0.194$,\\$\lambda_1=0.470$, $\lambda_2=0.504$\end{tabular}  \\
Chord HMM & \hspace{-3pt}\begin{tabular}{l}$\beta_1=0.94$, $\beta_2=4.70$, $\gamma_1=7.53$,\\$\gamma_2=5.29$, $\zeta=0.10$\end{tabular}\\
\bottomrule
\end{tabular}
}
\label{tab:OptimizedParameters}
\end{table}
The results show that the second-order and third-order HMMs significantly outperform other methods.
The clear difference between the match rates of the first-order and second-order HMMs indicate that incorporating longer-range dependence is indeed effective for improving the match rates.
On the other hand, although the third-order HMM consistently has higher match rates in all evaluation measures than the second-order HMM, the differences are small.
This indicates that the improvements for high-order HMMs seem to saturate at these orders, at least with the current amount of training data.
There is still a gap between the match rates of these current state-of-the-art methods and those among human players.
This point is further examined in the following sections.

The chord HMM, which is a probabilistic formulation of the model of \cite{Kasimi2007}, and the first-order HMM have similar match rates, but the latter consistently outperforms the former.
This is somewhat unexpected because the chord HMM incorporates transitional dependence of chords, which corresponds to long-range dependence at the note level, as well as the influence of note durations.
There are two possible reasons for this.
One is that the model architecture, particularly the decomposition of probabilities in Eqs.~(\ref{eq:ChordLevelHMMTrProb}) and (\ref{eq:ChordLevelHMMOutProb}), may be inaccurate.
The other is that the constraint for sustaining notes could possibly act in an unwanted way because the input piano performance data were synthetic data that contained unphysical note durations and temporal overlaps of notes.
Given this result, we have not pursued evaluation of higher-order chord HMMs.

The two DNN-based methods (FF and LSTM) have similar match rates and their general match rates are slightly lower than that of the first-order HMM.
Notably, their recombination match rates are significantly lower than that of the first-order HMM.
This indicates that the trained DNNs are less capable of capturing sequential consistency of fingering, which is incorporated as transition probabilities in the HMMs.
The result indicates that the problem of piano fingering estimation cannot be solved by a simple DNN-based method.
It should be noted that the performance of DNN-based methods is generally largely dependent on the amount of training data and the result does not preclude many possible refinements including the use of temporal features and data augmentation using the symmetric property of fingering data suggested in Secs.~\ref{sec:DescriptionOnChords} and \ref{sec:DesignOfOutProb}, respectively.
The LSTM network can be refined to a bi-directional network which can incorporate the information of future finger numbers.
Refinements of the optimization method are also possible, including the application of dropout and batch normalization.
Again, the purpose of this comparison is just to provide a reference. Further refinements of DNN-based methods are outside the scope of this study.

\begin{table}[t]
\centering
\caption{Comparative evaluation results for the VNS set. See the caption to Table \ref{tab:ComparativeEvaluation}.}
{\tabcolsep = 3pt
\begin{tabular}{rcccc}\toprule
Method & $M_{\rm gen}$ & $M_{\rm high}$ & $M_{\rm soft}$ & $M_{\rm rec}$\\
\midrule
1st HMM   & $60.5$ & $67.9$ & $82.4$ & $73.7$ \\
2nd HMM   & $63.1$ & $70.6$ & $85.1$ & $77.6$ \\
DNN (LSTM)& $59.8$ & $64.6$ & $82.6$ & $67.9$ \\
VNS \cite{Balliauw2017} & $56.7$ & $62.4$ & $78.2$ & $67.5$ \\
Human$^*$ & $70.2$ & $78.9$ & $91.0$ & $84.5$ \\
\bottomrule
\end{tabular}
}
\label{tab:ComparativeEvaluationVNS}
\end{table}
The VNS method was able to process only 14 pieces out of the 30 pieces in the test data (which is presumably because of failure in parsing the MusicXML files).
To compare the VNS method with other methods, the match rates on these 14 pieces (called the VNS set) were calculated and are compared with those of other methods in Table \ref{tab:ComparativeEvaluationVNS}, where to save space only results for representative HMM-based and DNN-based methods are shown.
We see that the match rates of the VNS method are consistently and significantly lower than those for the first-order HMM.
This shows the efficacy of the statistical learning approach based on HMMs.
In comparison with the LSTM, 
The VNS method has lower match rates than the LSTM, although the recombination match rates for the two methods are similar.

%%%
\subsection{Effect of the Architecture of HMMs}
%%%

%
\begin{table}[t]
\centering
\caption{Match rates for the first-order HMM with different architectures of output probabilities.}
{\tabcolsep = 3pt
\begin{tabular}{cccccc}\toprule
\begin{tabular}{c}Pitch\\representation\end{tabular} & \hspace{-3pt}\begin{tabular}{c}Chord\\constraint\end{tabular} & $M_{\rm gen}$ & $M_{\rm high}$ & $M_{\rm soft}$ & $M_{\rm rec}$\\
\midrule
Lattice  & $\checkmark$ & $61.6$ & $68.3$ & $82.8$ & $74.0$\\
Integral & $\checkmark$ & $61.4$ & $67.9$ & $82.6$ & $74.0$ \\
Lattice  &              & $60.0$ & $66.7$ & $81.2$ & $72.7$ \\
Integral &              & $59.5$ & $66.2$ & $80.6$ & $72.2$ \\
\bottomrule
\end{tabular}
}
\label{tab:EffectOfHMMArchitecture}
\end{table}
We also conducted comparative evaluations to examine the effect of the architecture of output probabilities of the HMMs.
The results listed in Table \ref{tab:EffectOfHMMArchitecture} were obtained in cases with $\alpha_1=1$ and only the transposition symmetry is imposed.
(We will discuss the effect of imposing additional symmetries in Section~\ref{sec:TrainDataSize}.)
We see that the constraint for chords consistently improves the match rates.
We can also see that the lattice pitch representation is better than the integral pitch representation, even though the improvements in the match rates are relatively small.
The results match our expectation because these refinements were intended to make the model more precise.

%%%
\subsection{Influence of Training Data Amount}
\label{sec:TrainDataSize}
%%%

\begin{figure}[t]
\centering
{\includegraphics[clip,width=0.6\columnwidth]{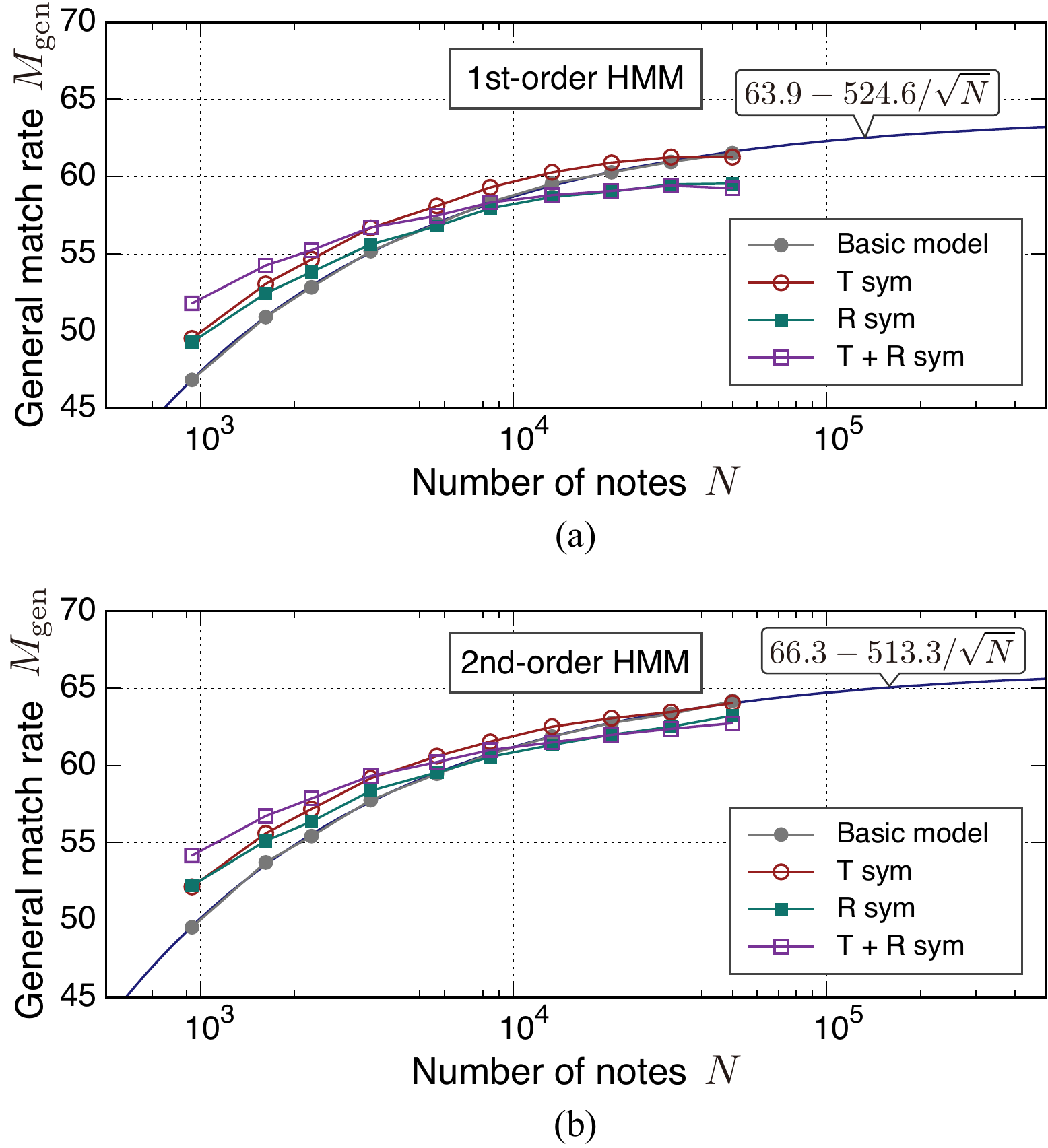}}
\caption{Variation of match rates for different amounts of training data. The basic model is the one without an additional symmetry, `T sym' indicates the time inversion symmetric model, `R sym' indicates the reflection symmetric model, and `T $+$ R sym' indicates the model with both the additional symmetries.}
\label{fig:TrainingDataScaling}
\end{figure}

We also investigated the influence of the amount of training data in order to confirm and predict the effect of increased training data and to examine the effects of imposing additional (time inversion and reflection) symmetries in the output probabilities of the HMMs.
In Fig.~\ref{fig:TrainingDataScaling}, we plot the general match rates for the first-order and second-order HMMs with varying amounts of training data.
In all cases we used lattice pitch representation and the sets of hyperparameters $\alpha_l$ and $\lambda_l$ listed in Table \ref{tab:OptimizedParameters}.
The rightmost points were obtained using the full training data, which correspond to the values in Table \ref{tab:ComparativeEvaluation}, and other points were obtained by training the models with reduced sets of data.
Specifically, we chose a number of pieces that is less than the whole training dataset, randomly sampled 100 different sets of pieces with this amount from the dataset, trained each model using these 100 datasets, and computed the average match rates for the test data.

The tendencies of the results are same for both the first-order and second-order HMMs.
When the whole training data is used, the basic model has the highest match rate and the model with both the additional symmetries has the lowest match rate; imposing the time inversion symmetry decreases the match rate slightly and imposing the reflection symmetry leads to a larger decrease.
This indicates that a model with fewer imposed symmetries, which has better descriptive power, has more predictive ability when it is trained with a sufficiently large amount of data.
It also indicates that the degree of asymmetry is larger for the reflection symmetry than the time inversion symmetry.
The ranking of the models changes with different amounts of training data: with the least amount the basic model has the lowest match rate, and imposing each of the additional symmetries increases the match rate.
This is because the additional symmetries reduce the effective number of model parameters and make the model more robust against overfitting, which is the main cause for deterioration of match rates in the limit of small training data.
The results indicate that the two symmetries are good approximations.

An interesting result is that the match rates can be fitted with the simple function
\begin{equation}
A(N)=a-b/\sqrt{N},
\label{eq:AsymptoticForm}
\end{equation}
where $N$ is the number of notes (sample size) in the training data and $a$ and $b$ are constants.
We show in \ref{app:AsymptoticForm} that this function can be derived from statistical theory when it is assumed that the match rate is a smooth function of the model parameters and the model parameter values are close to the asymptotic values.
An important consequence is that the coefficient $a$ gives the prediction for the limit of infinite training data, which can be seen in Fig.~\ref{fig:TrainingDataScaling}.
The result indicates both the potential and limitation of the models: we should be able to improve the match rate by about $2\%$ by further increasing the training data, but even with infinite training data the match rates of these models do not reach the degree of agreement among human players ($\sim71\%$).
It should also be noted that these predictions can be modified if the asymptotic values of model parameters change as we use more training data and that the match rate depends on test data.

%%%
\subsection{Example Results and Error Analysis}
%%%

%
\begin{figure}[t]
\centering
{\includegraphics[clip,width=0.99\columnwidth]{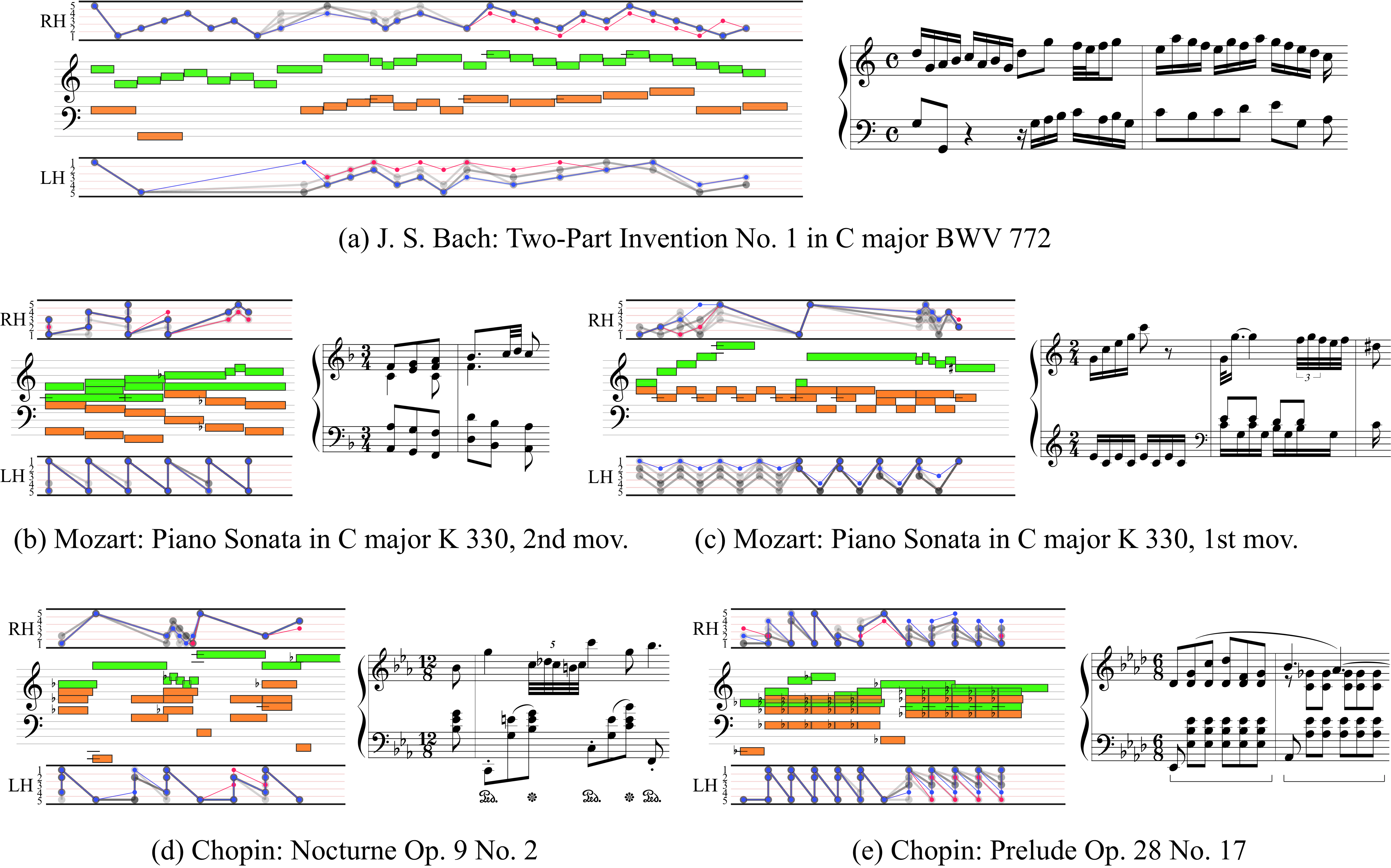}}
\caption{Example results of piano fingering estimation by HMMs. A piano roll representation, notated score, ground truth fingerings, and estimated fingerings are shown for each piece. In the piano roll representation, the green and orange rectangles represent notes in the RH and LH parts, respectively. The ground truth fingerings by multiple players are indicated with grey half-transparent circles and edges (a thicker color indicates a higher degree of match). The estimates generated by the first-order and second-order HMMs are indicated similarly in red and blue, respectively. Where these estimates are the same, only the fingering generated by the second-order HMM is shown.}
\label{fig:Ex}
\end{figure}
Here we discuss some example results to demonstrate the ability and limitation of the HMM-based methods as well as the effect of using a high-order model (Fig.~\ref{fig:Ex}).
In the RH part of the passage in Fig.~\ref{fig:Ex}(a), the estimates generated by the two models match in the first half and diverge after the note E5.
While both the estimated fingerings can be performed without difficulty, the second-order HMM matches better with the ground truths.
Indeed the fingering 4-2-5 for the G5-E5-A5 in the middle of the passage induces a smoother motion and the finger transition from G5 to A5 considered in the second-order HMM is important for the decision.
Similarly, the fingering for the LH part estimated by the first-order HMM has finger transitions that are acceptable when only each pair of consecutive notes are considered, but are globally much less common than those by the second-order HMM, which match better with the ground truths.
The importance of considering second-order finger transitions can also be understood from the results in Fig.~\ref{fig:Ex}(b), which shows a polyphonic example.
The RH fingering generated by the second-order HMM, which agrees with the ground truths, is preferred regarding the melody line (F4-G4-A4-B$\flat$4-C5-D5-C5), which involves high-order transitions at the note level.
As indicated by the relatively high match rates in Table \ref{tab:ComparativeEvaluation}, the fingerings generated by the HMMs are in most parts acceptable and show the models' nontrivial ability of sequence optimization\footnote{Readers can access the estimated results, which can be observed by the visualization tool used to create Fig.~\ref{fig:Ex}, from the accompanying webpage: http://statpianofingering.github.io/demo.html . The fingerings estimated by the second-order HMM for three pieces including the one in Figs.~\ref{fig:Ex}(a) and \ref{fig:Ex}(b) are also given there.}.

In the following, we focus on typical errors made by the HMM-based method\footnote{Here an `error' is formally defined as an estimated finger that is different from all of our ground truths. Since correct fingerings are not uniquely determined, as discussed earlier, such an `error' is not necessarily an impossible choice of finger.}.
In the LH part of Fig.~\ref{fig:Ex}(a), the erroneous fingering 1 for the third note G3 is caused by a phrase boundary.
Fingering transitions across phrases are less constrained than those inside phrases.
The finger transition 5-1 for G2-G3 (the second and third note onsets) is more common than transition 5-5 in general, but the latter is chosen by human players since it is considered as a dead interval separated by rests and finger 5 for the G3 is more proper for the later transitions.
A similar influence of phrasing leads to errors in the LH fingerings in Fig.~\ref{fig:Ex}(d).
In the triplet progressions, a bass note and two chords, both the first-order and second-order HMMs make errors at the second chord (at different positions).
The estimated fingerings might be better for the transition from the bass note to the second chord if they were slurred, but in fact the articulation induces focus on the transition from the second to third chords, which results in different fingerings.

In Fig.~\ref{fig:Ex}(c), the LH fingerings generated by the HMMs use a smaller range of fingers than that used by human players.
Similar cases are often found in the estimated fingerings.
As in the first half of the passage, isolated bichords of major/minor thirds that are normally played by finger pairs 1-3, 2-4, etc.\ are often assigned fingers 1-2, 2-3, etc.\ in both hand parts.
Some fifth intervals that are normally played by finger pair 1-5 are also sometimes assigned other finger pairs.
As the misassignment of finger 4 for the bass notes in the second half of the passage, upward/downward apex notes that are normally played by finger 1 or 5 are sometimes assigned other fingers.

In the RH part and the second half of the LH part in Fig.~\ref{fig:Ex}(e), it can be observed that the fingerings generated by the HMMs sometimes use wider ranges of fingers than those of human players.
In a passage like this one where the RH and LH overlap, smaller ranges of fingers than usual are often used, which is not considered in the HMMs where the two hands are modeled independently.
In addition, finger 5 assigned to the A$\flat$4 in the RH part by the HMMs is not appropriate as it hinders a smooth transition between the melody notes B$\flat$4-A$\flat$4.
As these notes are interrupted by chordal notes, long-range dependence should be taken into account to find a proper fingering.
Such very-long range dependence in the fingerings given by human players are not uncommon, and can lead to estimation errors.
There were also estimation errors induced by incorrect assignment of hand parts.

The above-described types of errors are related to high-level contexts of musical sequences that cannot be computationally described easily.
Although a quantitative error analysis was not possible, they are types often found.
This situation is even more complex as some estimation errors are induced by others because of the strong sequential dependence of fingering.
On the other hand, regardless of their frequencies, those typical errors demonstrate the limitation of the current models and give insight into how the models can be refined.

%%%
\subsection{Discussion}
%%%

While our results clearly demonstrated the potential of the statistical learning approach for piano fingering estimation, they also revealed limitations.
It should be noted that although individual cases of estimation errors can be avoided by heuristic rules, such an ad hoc treatment would not be a fundamental solution since fingering estimation is essentially a sequential optimization problem where the number of musical note combinations grows exponentially with respect to the sequence length.
Thus, it is necessary to systematically refine the model to improve the performance of fingering estimation.
Here we summarize possible extensions of the approach.

\paragraph{Temporal Features and High-Level Contexts} Temporal features such as inter-onset interval, duration, and the gap between the previous offset time and the current onset time are important to capture the irregularity of sequential dependence.
This irregularity includes the constraints specific to temporally overlapping notes and changes in the degree of freedom of finger choices.
The temporal features are also important for estimating the intended phrasing, which also influences the fingering.
High-level contexts to capture phrases, voices, apex notes, and repetitive structures also play relevant roles in finding optimal fingerings.
The chord HMM and DNNs may be important for solving these issues since the former can incorporate offset time information and chord-level long-range dependence and the latter are advantageous in dealing with many features in a computationally efficient way.

\paragraph{Interdependence of the Two Hands} Interdependence between fingerings of the two hands is important when the two hands overlap or when they have parallel motions \cite{Musafia1971,Parncutt1999}.
Estimation of hand parts, which is attempted in \cite{Nakamura2014}, is also necessary to fully understand piano fingering.

\paragraph{Individuality and Model Adaptation} While we have focused on the generic properties of piano fingering, individuality is also important.
Individual differences such as the lengths of fingers and the skill level are important factors of fingering decision \cite{Nagata2014,Sloboda1998}.
To enable adaptation to individual styles of fingering with minimal data, an ideal model represents these factors as variables and the generic properties as constants.
Fingering is also related to musical expressions and performance styles.
To capture this relation, it is necessary to consider additional musical score elements such as slurs and articulations, and additional performance elements including offset times, key press and release velocities, and pedalling, which were not studied here.

\paragraph{Cost vs.\ Statistical Viewpoints} The evaluation results showed the advantage of the statistical learning approach compared to the cost-based and constraint-based methods; many aspects of fingering decision can be explained by the principle of statistical learning and probabilistic optimization.
On the other hand, the results also showed its limitation.
As discussed in Section~\ref{sec:Intro}, to ultimately understand the mechanism of fingering, it is considered necessary to quantify fingering costs.
The small ranges of fingers used in the generated fingering in Fig.~\ref{fig:Ex}(c) exemplify the fact that high-probability fingerings and low-cost fingerings are different in general.
Fingering costs and probabilities are the cause and effect of the underlying decision process and the former cannot be simply inferred from the latter.
It should be noted, however, that statistical learning may have its own relevance in the fingering decision process, as indicated from the significant difference between the contemporary fingering practice (e.g.\ \cite{Musafia1971}) and the traditional practice (e.g.\ \cite{Couperin}).
Further investigations on these problems are left for future work.

%%%%%%%%%%%%%%%%%%%%%%%%%%%%%%%%%%%%%%%%%%%%%%
\section{Conclusion}
\label{sec:Concl}
%%%%%%%%%%%%%%%%%%%%%%%%%%%%%%%%%%%%%%%%%%%%%%

In this paper, we introduced a newly released dataset (PIG Dataset) and studied statistical learning and estimation methods for piano fingering.
The statistical modeling approach has the advantage of enabling efficient parameter estimation from data and even the first-order HMM outperformed the representative constraint-based method \cite{Balliauw2017} in terms of the estimation accuracies, when the model was trained with a sufficient amount of data.
Incorporating longer-range sequential dependence by means of higher-order HMMs further increased the accuracies and achieved the current state-of-the-art performance.
We also compared DNN-based methods with basic network architectures and found that the performances of the DNNs were worse than that of the HMMs.
The best estimation accuracy we achieved is however still lower than the degree of agreement among human players, and we discussed possible directions for refining the models.
The newly released dataset of piano fingering, the results of statistical analysis on the individual difference, and the proposed evaluation measures can be used in future studies.

The present generative model of piano fingering has applications other than fingering estimation.
For constructing more refined models/methods for fingering estimation, more training data is desired.
Automated methods for data collection using video processing \cite{MacRitchie2013,Oka2013}, optical music recognition \cite{Chen2016}, and/or censoring \cite{Randolph2016} would be necessary for extensively increasing the amount of data.
The present fingering model can be used as a prior model for complementing data exposed to noise and enhancing the precision of such automation techniques.
The model can also be used as a prior model of piano scores for music transcription and generation tasks \cite{Nakamura2018}, to control the performance difficulty and induce the output scores more similar to human-composed one.

% ------------------------------------------------------------------------------
% Appendix
% ------------------------------------------------------------------------------
\appendix

%%%%%%%%%%%%%%%%%%%%%%%%%%%%%%%%%%%%%%%%%%%%%%
\section{Algorithm for Computing the Recombination Match Rate}
\label{app:RecombMatchRate}
%%%%%%%%%%%%%%%%%%%%%%%%%%%%%%%%%%%%%%%%%%%%%%

In this appendix we explain the algorithm for computing the recombination match rate, which was introduced in Section~\ref{sec:EvaluationMeasure}.
An estimated sequence (of finger numbers) is denoted by $(f^{\rm est}_n)_{n=1}^N$ ($N$ is the number of notes) and multiple ground truths are denoted by $(f^{(g)}_n)_{n=1}^N$ $g\in\{1,\ldots,N_g\}$, where $N_g$ is the number of ground truths.
The recombination edit cost $E_{\rm rec}$ is formulated as the total cost to reconstruct the estimated sequence from the multiple ground truths.
The recombined ground truth $(f^{\rm rec}_n)_{n=1}^N$ can be represented as $f^{\rm rec}_n=f^{(z_n)}_n$, where $z_n\in\{1,\ldots,N_g\}$ indicates the reference ground truth at the $n$\,th note.

The recombination cost can be represented as transition cost for the sequence $z_n$ as
\begin{equation}
C(z_{n-1},z_n)=\begin{cases}
0, & z_n=z_{n-1};\\
C_{\rm rec}, &z_n\neq z_{n-1}\text{ and }f^{(z_n)}_n=f^{(z_{n-1})}_n;\\
C'_{\rm rec}, &{\rm otherwise}.\\
\end{cases}
\end{equation}
Here, $C_{\rm rec}$ is the recombined cost at those locations where the finger numbers of the recombined ground truths match, as described in Section~\ref{sec:EvaluationMeasure}, and we have also introduced recombined cost $C'_{\rm rec}$ at those locations where the finger numbers do not match.
This is a simple generalization and if we set $C'_{\rm rec}=\infty$ we recover the recombination cost defined in Section~\ref{sec:EvaluationMeasure}.
The substitution cost can be represented as a function of the recombined ground truth $f^{\rm rec}_n$ and the estimation $f^{\rm est}_n$ as
\begin{equation}
C(f^{\rm rec}_n,f^{\rm est}_n)=\begin{cases}
0, & f^{\rm rec}_n=f^{\rm est}_n;\\
C_{\rm sub}, & f^{\rm rec}_n\neq f^{\rm est}_n.\\
\end{cases}
\end{equation}
The recombination edit cost $E_{\rm rec}$ is then expressed as
\begin{equation}
E_{\rm rec}(z_{1:N})=\sum_{n=2}^NC(z_{n-1},z_n)+\sum_{n=1}^NC(f^{(z_n)}_n,f^{\rm est}_n).
\end{equation}

To compute the recombination match rate, we need to find the recombination ground truth (indicated by $z_{1:N}$) that minimizes $E_{\rm rec}(z_{1:N})$.
This can be done with a standard dynamic programming similar to the Viterbi algorithm.
After this optimal recombination $\hat{z}_{1:N}$ is obtained, the recombination match rate is computed as $M_{\rm rec}=(N-E_{\rm rec}(\hat{z}_{1:N}))/N$.

%%%%%%%%%%%%%%%%%%%%%%%%%%%%%%%%%%%%%%%%%%%%%%
\section{Asymptotic Form of Match Rates}
\label{app:AsymptoticForm}
%%%%%%%%%%%%%%%%%%%%%%%%%%%%%%%%%%%%%%%%%%%%%%

In this appendix we derive Eq.~(\ref{eq:AsymptoticForm}), which gives the asymptotic form of match rates.
Suppose that we have a statistical model with parameters $\bm\theta$.
In the HMMs considered in the main text, $\bm\theta$ denotes the set of transition and output probabilities.
We first show that in the limit of a large number $N$ of samples the following asymptotic formula holds:
\begin{equation}
\bm\theta\sim\bm\theta_0+\frac{\bm f(\bm\theta_0)}{\sqrt{N}}+{\cal O}\bigg(\frac{1}{N}\bigg),
\label{eq:AsympFormParam}
\end{equation}
where $\bm\theta_0$ is the asymptotic value of $\bm\theta$ and $\bm f$ is some function that has the same dimensionality as $\bm\theta$.
This is a consequence of the Cram\'{e}r-Rao bound for an unbiased estimator \cite{CoverThomas}
\begin{equation}
\langle(\bm\theta-\bm\theta_0)^2\rangle\geq I(\bm\theta_0)^{-1}/N,
\label{eq:CramerRaoBound}
\end{equation}
where the LH side is the variance of $\bm\theta$, which is interpreted as the squared statistical error of $\bm\theta$, and $I(\bm\theta_0)$ is the one-sample Fisher information.
For a wide class of maximum-likelihood estimators including those for the discrete distributions used in this paper, it is known that the lower bound is achieved asymptotically \cite{Lehmann}.
One can see that the equality of (\ref{eq:CramerRaoBound}) is equivalent to Eq.~(\ref{eq:AsympFormParam}) by noting that $\bm\theta\to\bm\theta_0$ as $N\to\infty$.

Let us now consider a smooth function $A(\bm\theta)$ of statistics $\bm\theta$.
Substituting Eq.~(\ref{eq:AsympFormParam}) and taking the first-order Taylor expansion, we have
\begin{equation}
A(\bm\theta)\sim A(\bm\theta_0)+\frac{\partial A(\bm\theta_0)}{\partial\bm\theta_0}\cdot\frac{\bm f(\bm\theta_0)}{\sqrt{N}}+{\cal O}\bigg(\frac{1}{N}\bigg).
\end{equation}
This is equivalent to Eq.~(\ref{eq:AsymptoticForm}) in the large $N$ limit when one fixes $N$ and takes the expectation with respect to $\bm\theta$.

We see from the above derivation that Eq.~(\ref{eq:AsymptoticForm}) holds if the match rate is a smooth function of the model parameters $\bm\theta$ in the neighborhood of their asymptotic values $\bm\theta_0$.
It should be noted that the constants $a$ and $b$ in Eq.~(\ref{eq:AsymptoticForm}) are functions of the asymptotic value $\bm\theta_0$; their values change depending on the asymptotic value of $\bm\theta$.

%%%%%%%%%%%%%%%%%%%%%%%%%%%%%%%%%%%%%%%%%%%%%%
\section*{Acknowledgment}
%%%%%%%%%%%%%%%%%%%%%%%%%%%%%%%%%%%%%%%%%%%%%%

We thank Shigeki Sagayama and Kenji Watanabe for useful discussions, Shinichi Furuya and others for helping us collect fingering data, Dorien Herremans for providing the source code for the VNS method, and Sei Ueno for running the DNN-based methods.

\end{document}